\documentclass[runningheads]{llncs}
\usepackage[T1]{fontenc}
\usepackage{orcidlink}
\usepackage[sort&compress,numbers]{natbib}
\usepackage{graphicx}
\begin{document}
\title{Sentiment Analysis on the young people's perception about the mobile Internet costs in Senegal}
\titlerunning{Mobile Internet Sentiment Analysis}

\author{Derguene Mbaye\inst{1}\orcidlink{0000-0002-7490-2731} \and
Madoune Robert Seye\inst{1} \and Moussa Diallo\inst{1} \and Mamadou Lamine Ndiaye\inst{1} \and Djiby Sow\inst{2} \and Dimitri Samuel Adjanohoun\inst{2} \and Tatiana Mbengue\inst{2} \and Cheikh Samba Wade\inst{2} \and De Roulet Pablo\inst{3} \and Jean-Claude  Baraka Munyaka\inst{3} \and Jerome Chenal\inst{3}
}
\authorrunning{D. Mbaye et al.}
%

\institute{Polytechnic School (ESP), Dakar, Senegal\\ \email{derguenembaye@esp.sn}\inst{1}, \email{robertseye@hotmail.fr}\inst{1},
\email{moussa.diallo@ucad.edu.sn}\inst{1}\\ \and
Gaston Berger University (UGB) (Saint louis, Senegal) \and
Federal Institute of Technology Lausanne (EPFL) (Lausanne, Switzerland)
}

\maketitle
\begin{abstract}
Internet penetration rates in Africa are rising steadily, and mobile Internet is getting an even bigger boost with the availability of smartphones. Young people are increasingly using the Internet, especially social networks, and Senegal is no exception to this revolution. Social networks have become the main means of expression for young people. Despite this evolution in Internet access, there are few operators on the market, which limits the alternatives available in terms of value for money. In this paper, we will look at how young people feel about the price of mobile Internet in Senegal, in relation to the perceived quality of the service, through their comments on social networks. We scanned a set of Twitter and Facebook comments related to the subject and applied a sentiment analysis model to gather their general feelings.
\end{abstract}

\keywords{Sentiment analysis  \and Social media \and Social Web \and Language Models \and Low-resource \and African languages \and Low-resource languages \and Wolof.}

\section{Introduction}
Social networking has taken off in leaps and bounds around the world, and Africa is no exception. People share their opinions on all kinds of subjects, share achievements in their lives or simply engage in chit-chat. In $2024$, the Digital Global Overview Report\footnote{\url{https://datareportal.com/reports/digital-2024-global-overview-report}} recorded more than 5 billion active users on social networks, representing $62.3\%$ of the world's population. In Senegal, $20.6\%$\footnote{Social media users may not represent unique individuals} of the population is on social networks, according to the same report. With a median age of around $18$, this is a particularly young population, suggesting that they represent the vast majority of social network users. \texttt{Facebook}, \texttt{Twitter} and, more recently, \texttt{Tiktok} are the flagship platforms most used by the population. However, despite a growing mobile Internet penetration rate in Africa (one of the key factors in the rise of social networks in Africa), only two countries (South Africa and Mauritius) have achieved the 'advanced' status in the 2023 GSMA Connectivity Index\footnote{\url{https://www.mobileconnectivityindex.com/index.html\#year=2023}}.
The report shows that $42\%$ of adults in low-income countries are still not using mobile internet, despite being covered by a mobile broadband network. Several factors were identified, including a lack of the necessary knowledge and skills, and the inability to afford an internet-connected phone, data plans and other service fees. To emphasize the latter, it is common to find posts on social media in Senegal about the cost of mobile internet, and end-users' perception of network quality, among other things. \texttt{Twitter} in particular is an ideal source because of its audience, the variety of its users and its micro-blogging nature \cite{pak-paroubek-2010-twitter} facilitating the sharing of opinions through short messages\footnote{Twitter offers longer messages since 2023 through its premium feature, but short messages are still the favourite format among users.}. 

\begin{figure}[htbp]
\centerline{\includegraphics[width=1\textwidth, keepaspectratio]{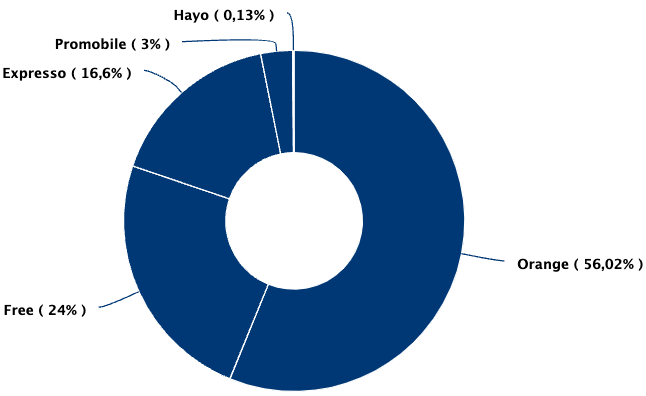}}
\caption{Operator market shares as of March 31, 2024, published by the Senegalese Telecommunications Regulatory Agency (ARTP).}
\label{telco-market}
\end{figure}

There are five (05) operators active in Senegal, including \texttt{Orange}, which has the largest market share, as shown in Fig. \ref{telco-market}.

\texttt{Orange} is the brand name of \texttt{SONATEL}, the country's incumbent operator, with \texttt{France Telecom}  as its majority shareholder since its privatization in 1997. It therefore occupies a dominant position, having own most of the infrastructure in place. Although it has the highest number of active users, it is not uncommon to see negative reviews on social networks about the cost of accessing its services, suggesting inequalities in coverage and quality of service between operators.

In this article, we study the opinions of young people on the cost of the mobile Internet in Senegal, using social networking platforms such as \texttt{Twitter} and \texttt{Facebook}. We show how a corpus can be built up through these platforms and how it can be used as a pressure tactic on telecoms operators. We collected a corpus of more than $10.000$ text posts distributed between three types of sentiments:
\begin{enumerate}
  \item texts containing \texttt{positive} emotions, such as happiness, amusement or joy ;
  \item texts containing \texttt{negative} emotions, such as sadness, anger or disappointment ;
  \item objective or \texttt{neutral} texts that only state a fact or do not express any emotions.
\end{enumerate}

We perform a linguistic analysis of our corpus and use a multilingual language model (LM) as a sentiment classifier to illustrate users' feelings. The paper is therefore structured as follows:
\begin{itemize}
  \item We begin by presenting some work done on sentiment analysis applied to various fields, including telecoms in Section \ref{rel-work}.
  \item Our data collection approach is presented in Section \ref{data}.
  \item Analysis of collected data is performed in section \ref{analysis}.
  \item In Section \ref{sentiment}, we present an analysis of the extracted sentiments and the approach adopted.
  \item We present some limitations of our methodology in section \ref{limit}
  \item Conclusion and perspectives are presented in section \ref{concl}.
\end{itemize}

\section{Related Work}\label{rel-work}
Opinions on social networks have been the subject of many studies in the literature, on subjects ranging from politics to health issues and natural disasters, among others. Researchers in \cite{pak-paroubek-2010-twitter} propose an efficient way to collect text data from Twitter for sentiment analysis and opinion mining. Their method allows automatic text collection based on sentiment (positive or negative) in such a way that human intervention is not required for classification. They worked on the English language, although the method is reusable in other languages. A sentiment analysis benchmark on African languages was proposed in \cite{muhammad-etal-2023-afrisenti} covering $14$ languages from $04$ different families with data sourced from Twitter. Tweets were manually labeled by native speakers, highlighting the challenges of working with African languages. Regarding the telecommunications field, the customer feedback and review on mobile telecommunication services in Malaysia has been studied in \cite{TelcoSentiment} using a Naïve Bayes sentiment analysis approach on Twitter data. Researchers in \cite{10010357} studied user complaints about internet quality during the COVID-19 pandemic in Indonesia with a CNN-based classifier to classify feelings about telecom operators. Data were collected on Twitter and underwent preprocessing and weighting of Word2Vec embeddings. To enhance the quality of service provided by Mobile Phone operators working in Pakistan, Twitter data has been analyzed in \cite{Saleem_Jamil_Mehmood_2023} to perform sentiment analysis in order to enable organizations to gain better insights regarding quality-of-service improvement. A framework has been proposed in \cite{9653423}, built on top of the Hadoop ecosystem, for analyzing data from Twitter using a domain-specific Lexicon in Greek. 

Twitter has been extensively studied in the literature, but due to the limitations imposed since its takeover by Elon Musk, it has become extremely restrictive to limit oneself to this platform\footnote{\href{https://datareportal.com/reports/digital-2023-deep-dive-the-state-of-twitter-in-april-2023}{What’s really going on with Twitter? - Data Reportal}}. Since then, hundreds of research projects have been cancelled, halted, or pivoted to other platforms as a result of these changes\footnote{\href{https://www.cjr.org/tow_center/qa-what-happened-to-academic-research-on-twitter.php}{Q\&A: What happened to academic research on Twitter? - CJR}}, and a significant decline in the commitment of researchers has been noted \cite{Bisbee_Munger_2024}. Facebook is a viable alternative, and similar work to that done on Twitter has been carried out there. Researchers in \cite{PMID:34764585} used Facebook for tracking the evolution of COVID-19 related trends. They collected a multilingual corpus covered 07 languages (English, Arabic, Spanish, Italian, German, French and Japanese)
and proposed an exhaustive analytics process including data gathering, preprocessing, LDA-based topic modeling and a presentation module using graph structure. A case study on text mining for Facebook and Twitter unstructured data analysis has been conducted in \cite{bank-analysis-nigeria} to help financial institutions in Nigeria analyse their competitor's social media sites and improve their decision-making. A sentiment analysis of Facebook comments was carried out in \cite{Syahriani_2020} concerning the presidential election in Indonesia in 2014. The authors targeted two official accounts among those of the candidates and used a Naïve Bayes classifier for sentiment analysis. A similar study was carried out in \cite{mexico} on the June 2017 local government campaign in the central state of Mexico. Researchers collected nearly 4,500 Facebook posts and performed a sentiment analysis on the text including emoticons raising a surprising paradox between perceived sentiment towards candidates and the actual election outcome. The perception of UBER\footnote{American multinational transportation company that provides ride-hailing services, courier services, food delivery, and freight transport.} users has been studied in \cite{uber} on the basis of Facebook posts published between July 2016 and July 2017. Corpus collection and sentiment analysis were carried out using a proprietary tool relying on a lexicon-based approach to categorizing sentiment. Using Facebook and Twitter accounts of the top three telecommunication companies in Ghana, researchers in \cite{data14} reveal insights from unstructured texts. They studied the customers feelings about operators products or brand using a lexicon-based approach. The opinions of the Internet service in Sudan have been studied in \cite{sudan} in the Arabic language and the Sudanese dialect, which is a low-resource setting. They applied an SVM classifier and another one based on Naïve Bayes on a corpus of a thousand Facebook comments.

\section{Data Collection}\label{data}

\begin{figure}[htbp]
\centerline{\includegraphics[width=1\textwidth, keepaspectratio]{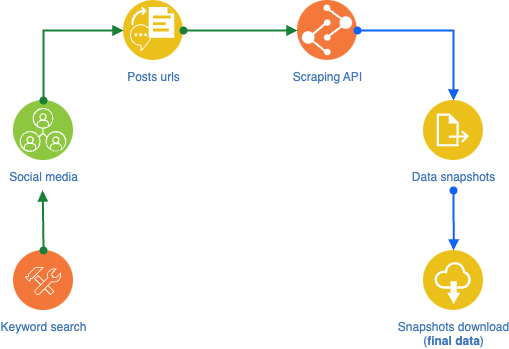}}
\caption{Diagram of the scraping strategy.}
\label{scraping-diag}
\end{figure}

An important characteristic of Twitter is its real-time nature. For example, when a disaster occurs, people make many Twitter posts (tweets) related to it, which enables its following easier simply by observing the tweets. However, due to limitations within the twitter platform, we were only able to collect a very limited extract of \texttt{250 tweets}. These tweets date back to a window of one week from the date of collection (September 23, 2024), and concern comments on a post made by the Orange operator on the lowering of mobile internet package prices. These types of posts are very rich in information, as users naturally bounce off them to share their opinions on package costs and how they feel about various aspects of the operator. We therefore adopted a similar approach to collecting information on Facebook, by targeting operators' posts on their Internet packages and then collecting the corresponding comments. To do this, we first performed a keyword search to identify these posts, collected the urls to them and used Bright Data's\footnote{\url{https://brightdata.com/}} scraping API to retrieve the corresponding snapshots. These snapshots represent all the data collected from the urls provided, which we then downloaded and stored in \texttt{csv} format. This last step was skipped for the Twitter scraping, where we used the Twitter API directly to download the data without using snapshots. We thus collected more than \texttt{10,000} Facebook and Twitter comments spread over 4 operators. The overall approach is illustrated in Fig. \ref{scraping-diag}.

We then proceeded to a data pre-processing stage in which we used regular expressions to remove all the hashtags, usernames and links of collected tweets. We also removed stop words, images and unnecessary columns returned during scraping. Stop words are a set of commonly used words in a language (e.g “a”, “the”, “is”...) used in Text Mining and Natural Language Processing (NLP) to eliminate words that are so widely used that they carry very little useful information. Stopwords lexicons for resource-rich languages like English are easily accessible on the Internet, but not for a language like Wolof (present in the collected data), making pre-processing difficult. Moreover, Wolof is written on social media in a way that differs from standard writing as illustrated in \cite{mbaye2023beqi}. This phenomenon results from the fact that Wolof is not taught at school, due to the fact that French has been adopted as the official language since colonization. As a result, people don't have a good grasp of its script, and tend to write it without regard to any writing rules, even though the language's alphabet is established. This difference in script tends to reduce the performance of existing language processing tools for Wolof, which are generally designed for the standard script.

\section{Data Analysis}\label{analysis}
\begin{figure}[htbp]
\centerline{\includegraphics[width=1\textwidth, keepaspectratio]{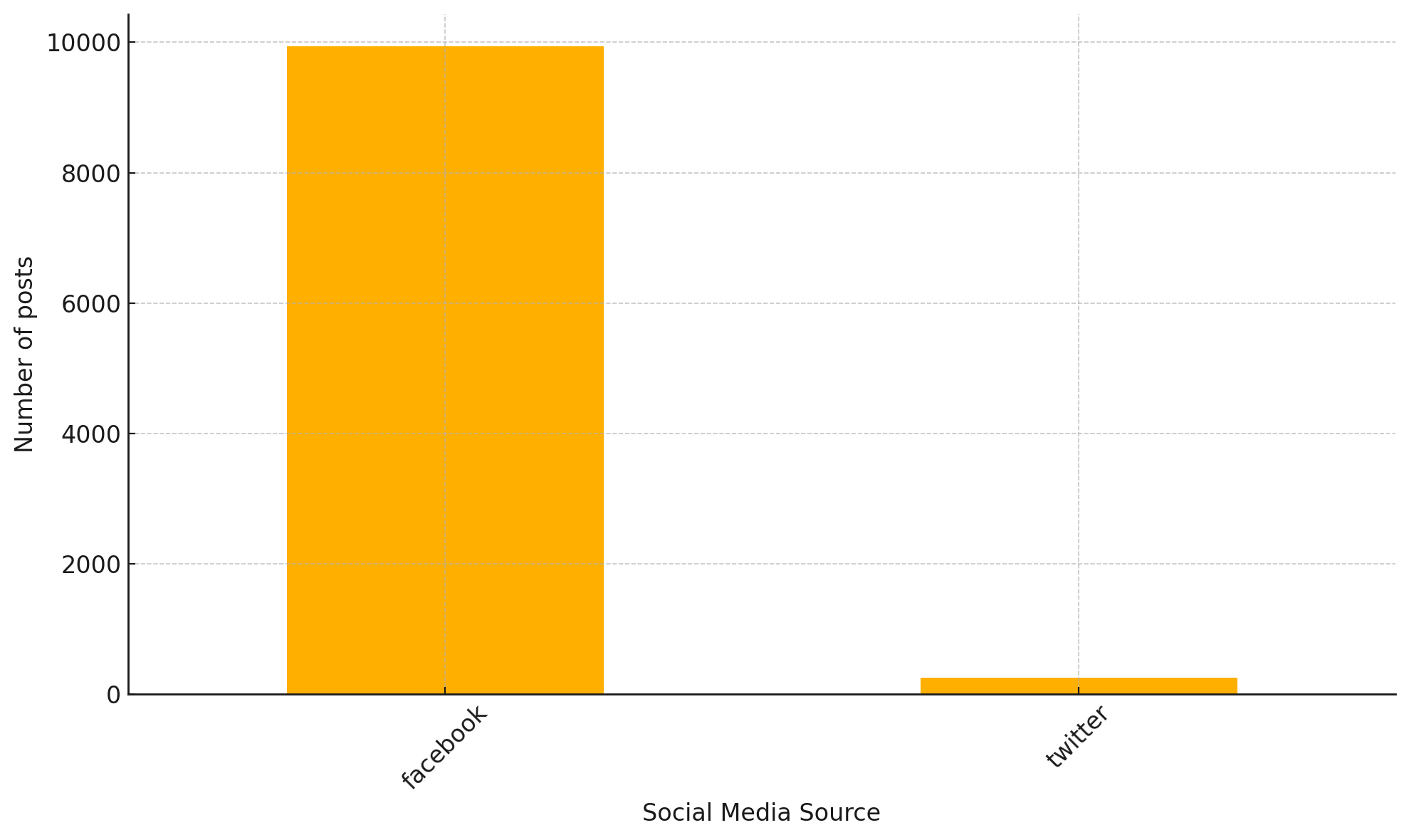}}
\caption{Distribution of Facebook versus Twitter data.}
\label{fbvstwitter}
\end{figure}

We were able to collect a total of \texttt{10184 posts}, originating mainly from Facebook, as illustrated in Fig. \ref{fbvstwitter}.

\begin{figure}[htbp]
\centerline{\includegraphics[width=1\textwidth, keepaspectratio]{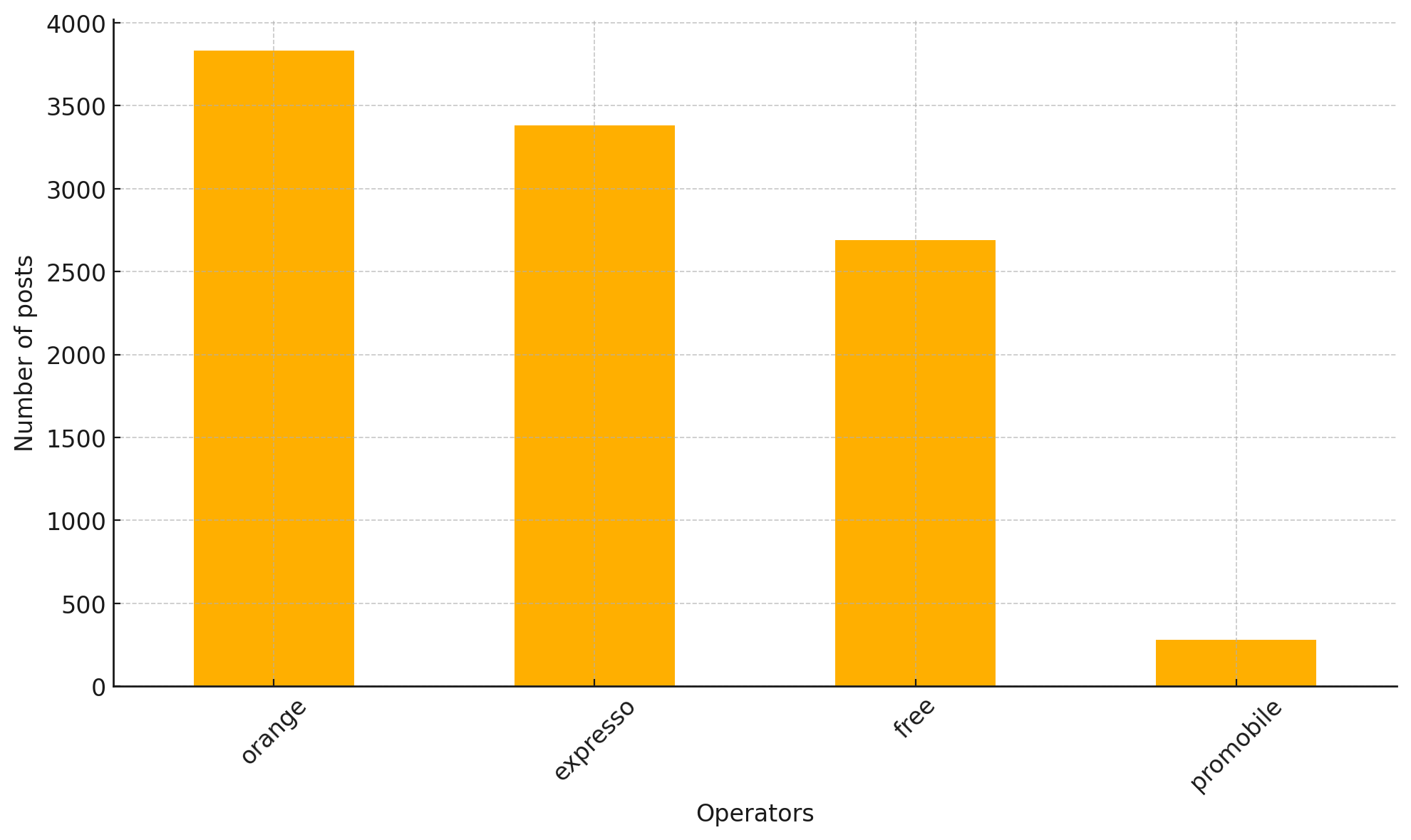}}
\caption{Data distribution across Senegal's 04 leading telecom operators.}
\label{datatelco}
\end{figure}

The data has been collected from the 04 main operators in Senegal, and the distribution of their data is shown in Fig. \ref{datatelco}.

\begin{figure}[htbp]
\centerline{\includegraphics[width=1.2\textwidth, keepaspectratio]{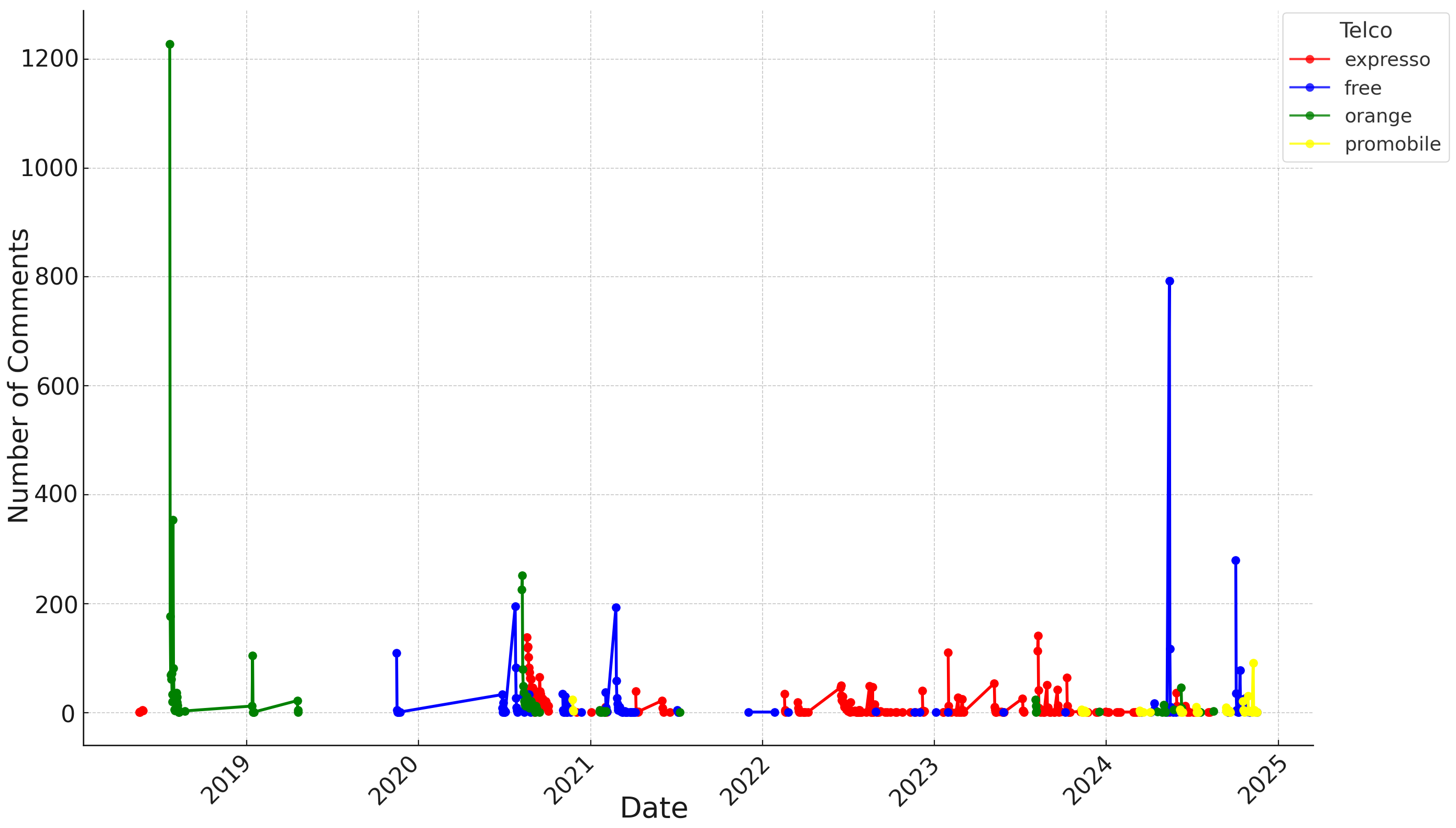}}
\caption{Data distribution across Senegal's 04 leading telecom operators.}
\label{period}
\end{figure}

The scraped data range from 2019 to 2024 for Facebook data and around September 2024 for Twitter data. Our access level to the Twitter API did not allow us to obtain the publication dates of the comments, therefore only those from Facebook are shown in Fig. \ref{period}. It's worth noting that the majority of comments relating to Orange were posted around 2019, while those for Free (its main competitor) were posted between 2021 and 2024. Promobile has the most recent comments (2024), while Expresso's comments are concentrated between 2023 and 2024. The period is therefore very important to consider, as external circumstances can influence user sentiment towards a particular operator.

\begin{figure}[htbp]
\centerline{\includegraphics[width=1\textwidth, keepaspectratio]{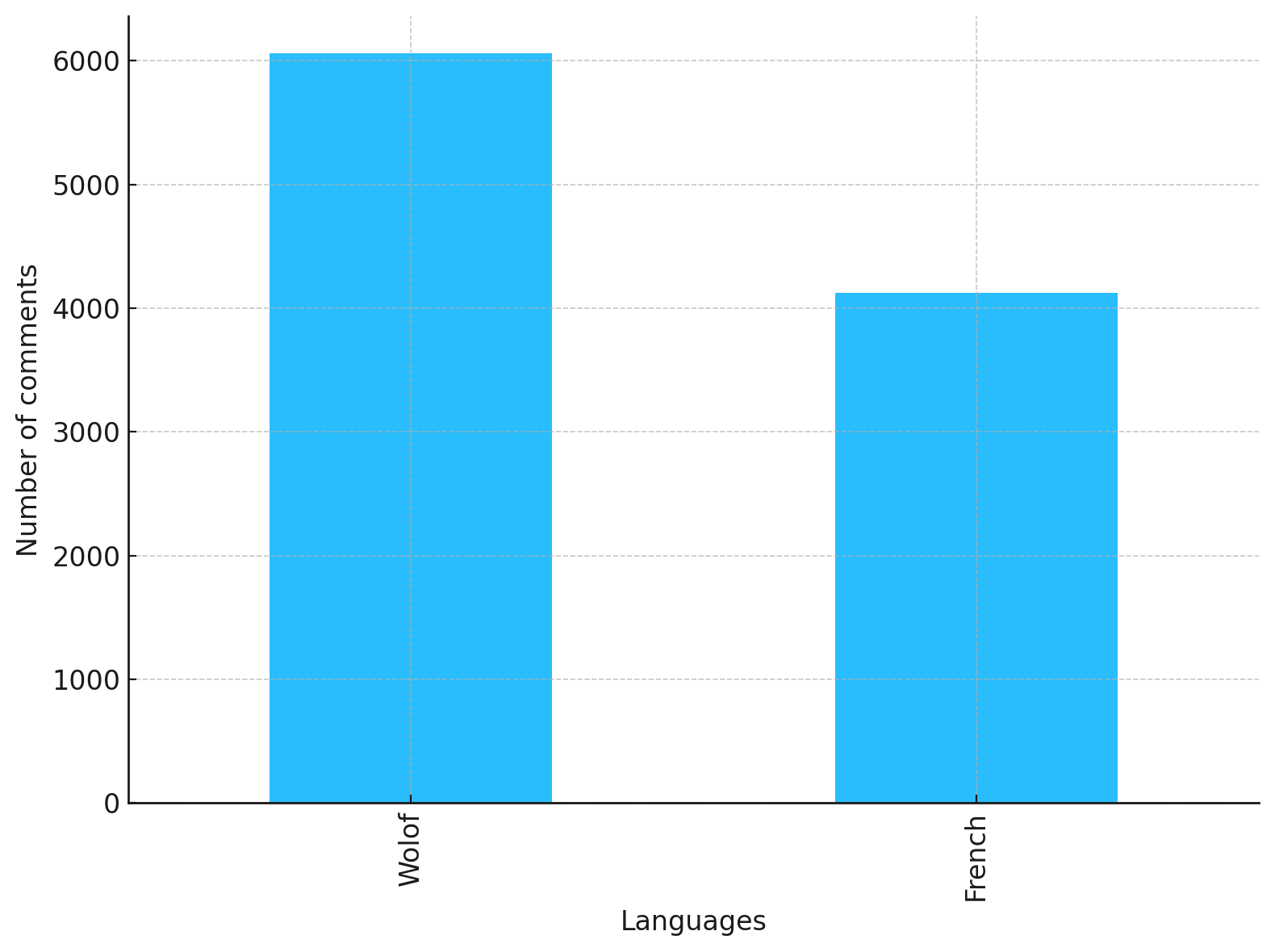}}
\caption{Proportion of comments in French compared to those in Wolof.}
\label{lang}
\end{figure}

\begin{figure}[htbp]
\centerline{\includegraphics[width=1\textwidth, keepaspectratio]{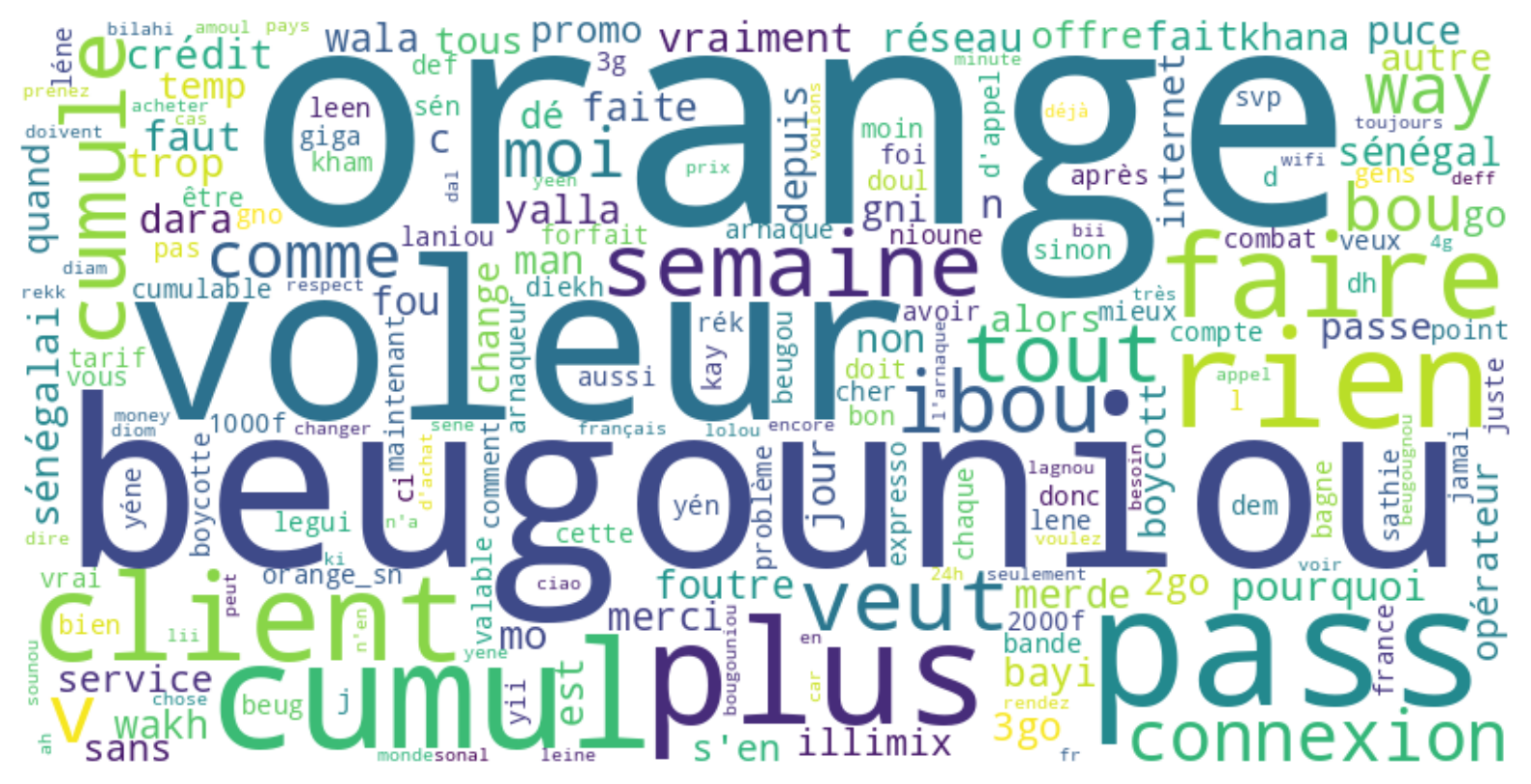}}
\caption{Word cloud of the most frequent words used in the comments from the Orange operator Senegal.}
\label{orange_cloud}
\end{figure}

With two dominant languages, French (the official language) and Wolof (the lingua franca), it's not uncommon to see comments in both languages on social networks. People also tend to mix the two in the same speech, a phenomenon known as code mixing or code switching \cite{surveycodeswitchedspeechlanguage}, typical of countries where several languages are present. To identify the languages of the collected comments, we used the GlotLID library \cite{kargaran-etal-2023-glotlid}, capable of identifying over 1,500 languages, including a wide coverage of low-resource ones. We were thus able to identify a higher proportion of texts in Wolof compared to those in French, as shown in Fig. \ref{lang}.
For this reason, it is essential to take both languages into account when pre-processing data.
Since we are mostly dealing with text data, it is quite handy to take a look at the list of all the possible words that were present in our posts (Twitter and Facebook comments). A useful concept called \texttt{word cloud} can be used for this task of representing the presence of various words in our list of comments. The size of the words represents the frequency of occurrence and we can quickly have an overview of the most used words based on this information. Given that stop words are very frequent but not significant, keeping them will tend to "spam" the word clouds. To remove them, we used the Natural Language Toolkit (NLTK) \cite{loper2002nltknaturallanguagetoolkit}, which is a popular suite of libraries and programs for symbolic and statistical natural language processing (NLP). It natively integrates French stop words, but not those in Wolof. To mitigate this constraint, we manually translated some French stop words into Wolof and then generated word clouds, identified residual stop words and added them to those in NLTK to clean up the final word clouds as much as possible. We finally generated word clouds on the comments from each operator and can observe an interesting phenomenon from here in terms of the words used for Orange compared to its competitors. In Fig. \ref{orange_cloud}, we see very hostile terms like \texttt{voleur} (thief), \texttt{arnaque} (scam), \texttt{boycotte} (boycott) or \texttt{beugouniou} (a Wolof term meaning "we don't want" or "we don't want it"). Indeed, Orange is generally criticized by the population for the high cost of its packages and the speed with which users perceive the consumption of their plans. As a result, they feel "robbed" and often call for a boycott.

\begin{figure}[htbp]
\centerline{\includegraphics[width=1\textwidth, keepaspectratio]{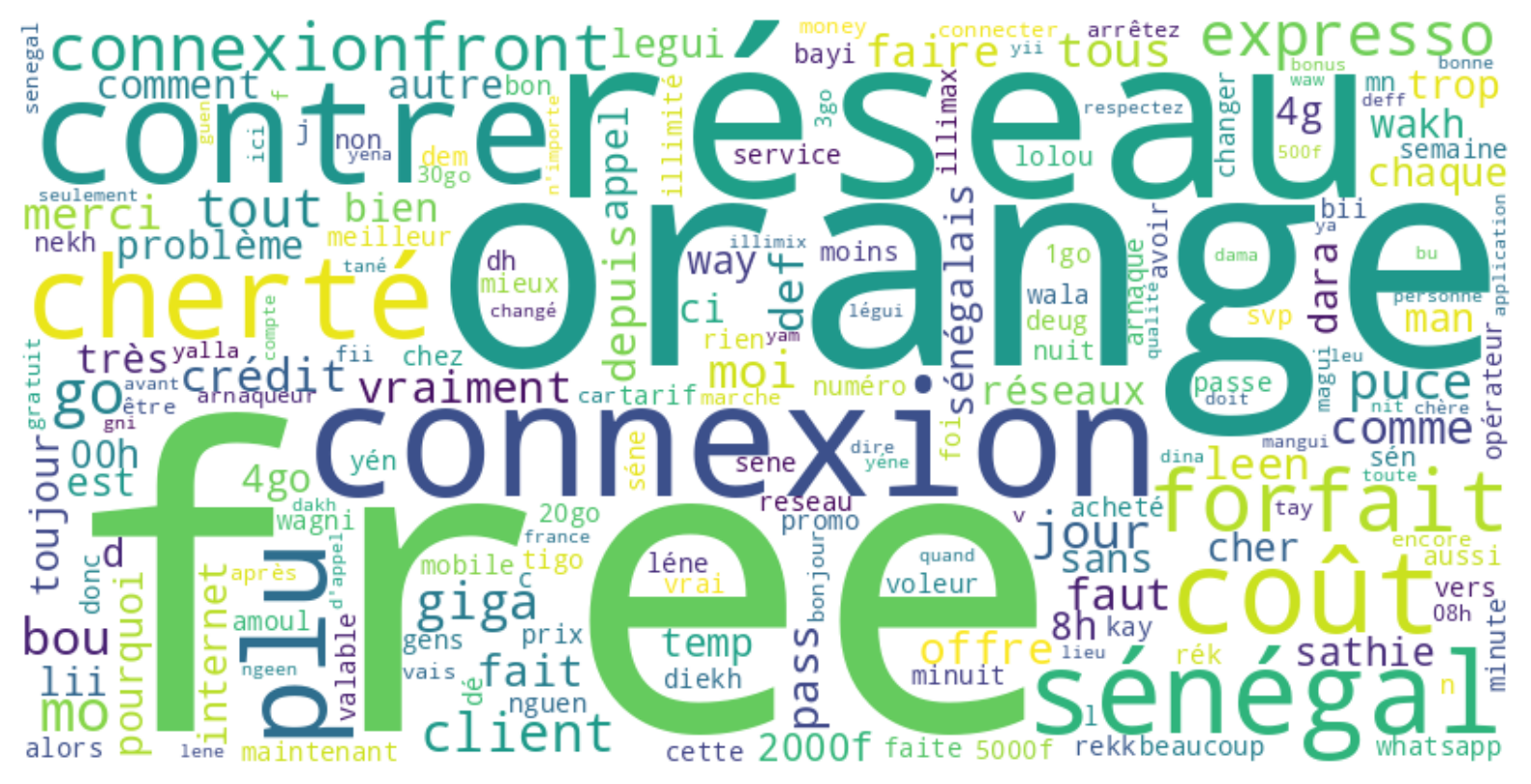}}
\caption{Word cloud of the most frequent words used in the comments from the Free operator Senegal.}
\label{free_cloud}
\end{figure}

\begin{figure}[htbp]
\centerline{\includegraphics[width=1\textwidth, keepaspectratio]{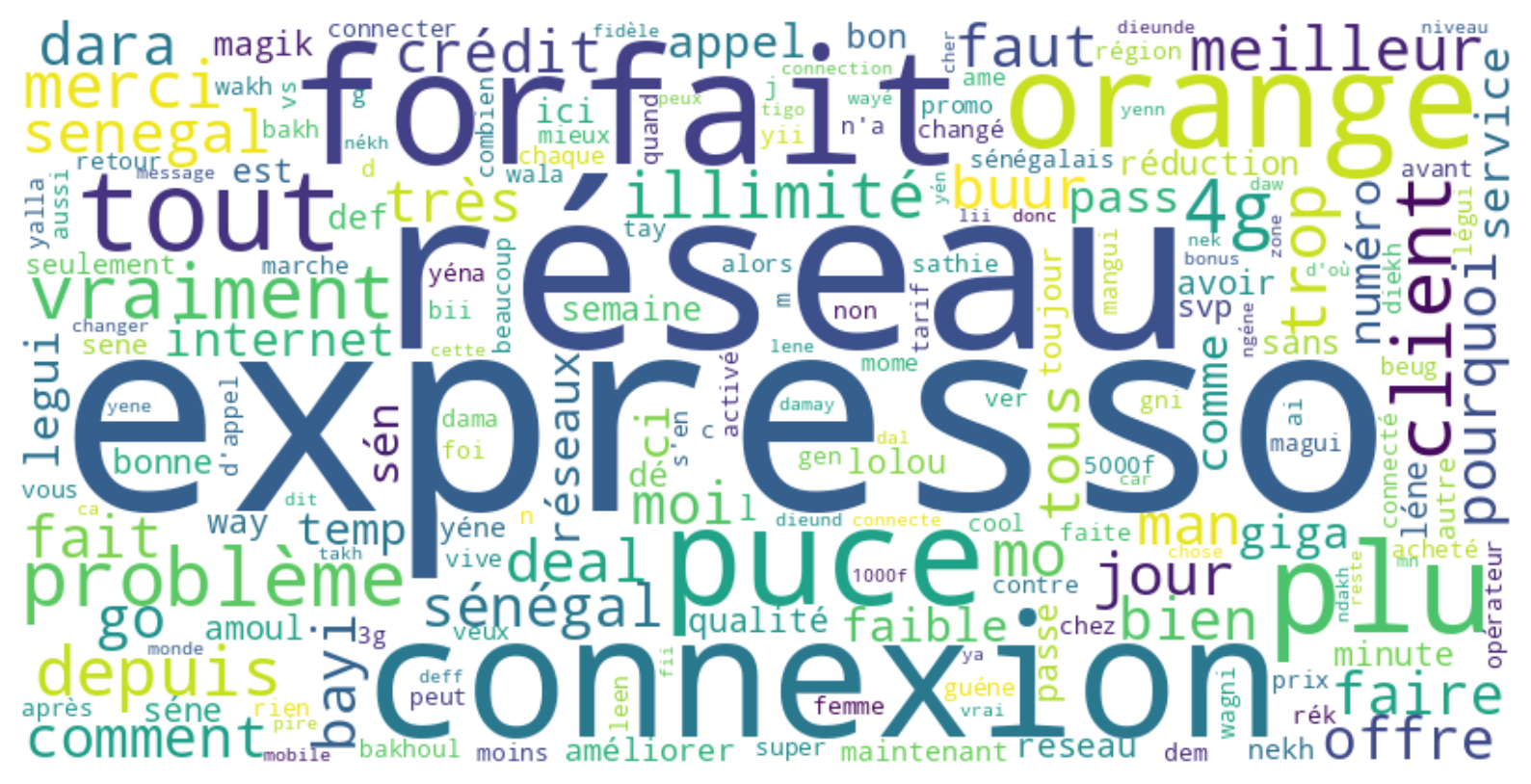}}
\caption{Word cloud of the most frequent words used in the comments from the Expresso operator Senegal.}
\label{expresso_cloud}
\end{figure}

\begin{figure}[htbp]
\centerline{\includegraphics[width=1\textwidth, keepaspectratio]{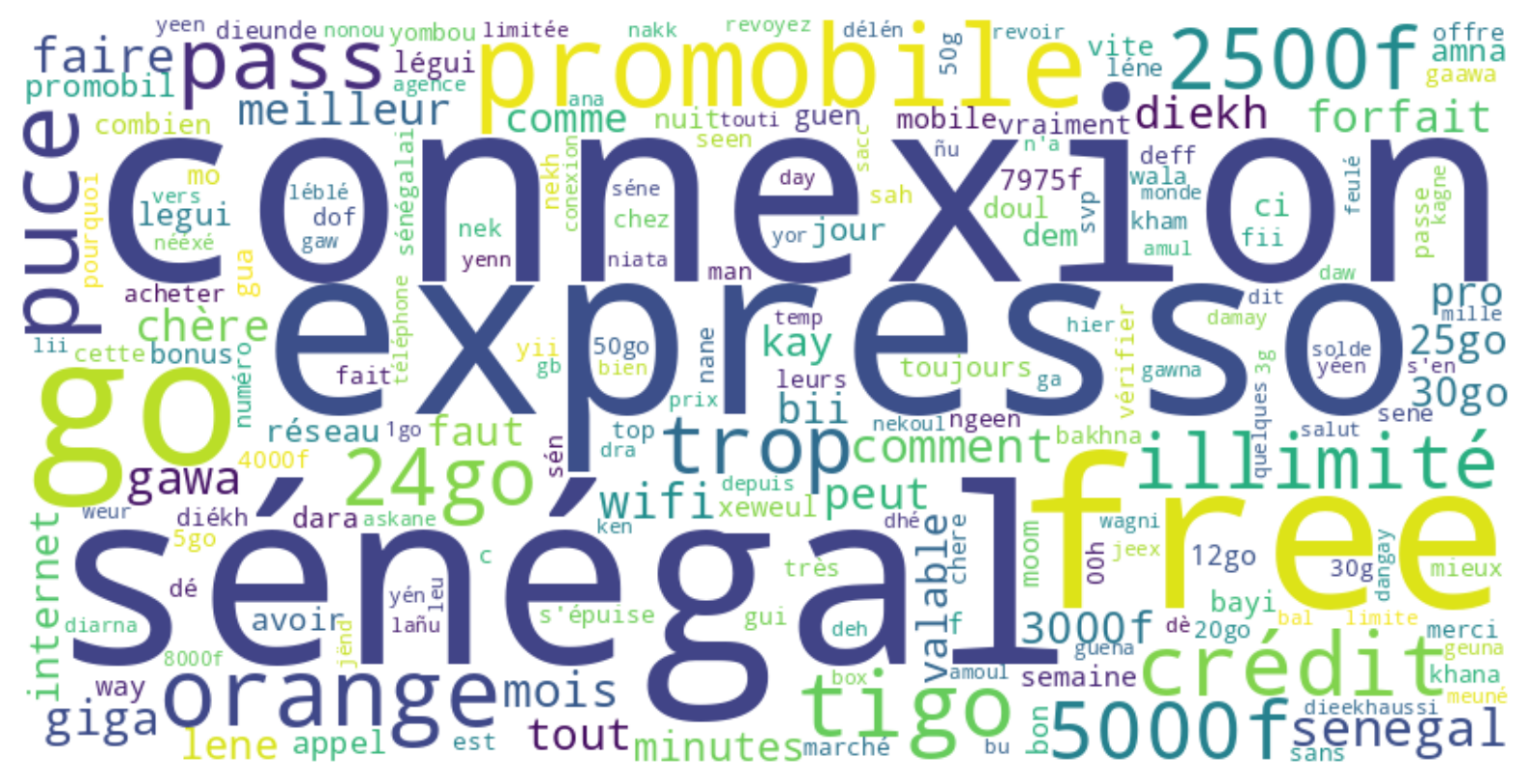}}
\caption{Word cloud of the most frequent words used in the comments from the Promobile operator Senegal.}
\label{promobile_cloud}
\end{figure}

In Free's word cloud in Fig. \ref{free_cloud}, terms like \texttt{front}, \texttt{contre} (against), \texttt{cherté} (expensive) and \texttt{coût} (cost) appear, illustrating Free's position as the operator who fights against the high cost of packages. Free is seen as a viable alternative to Orange, as is Expresso, but both operators are often criticized for the lower quality of their networks. For this reason, we see terms like \texttt{réseau} (network), \texttt{connexion} (connection), \texttt{problème} (problem) in their word clouds, especially in that of Expresso presented in Fig. \ref{expresso_cloud}. We note a similar trend for promobile illustrated in Fig. \ref{promobile_cloud}, a virtual operator based on the Orange network, singled out for its potentially expensive packages and still very poor network coverage. The trends observed in this disparity partly explain the dominance of Orange, despite the hostile terms often noted in user comments. Despite calls for boycotts, users find it hard to switch to a competitor, as the latter often offers inferior network quality and coverage.


\section{Sentiment Analysis}\label{sentiment}
\begin{figure}[htbp]
\centerline{\includegraphics[width=1\textwidth, keepaspectratio]{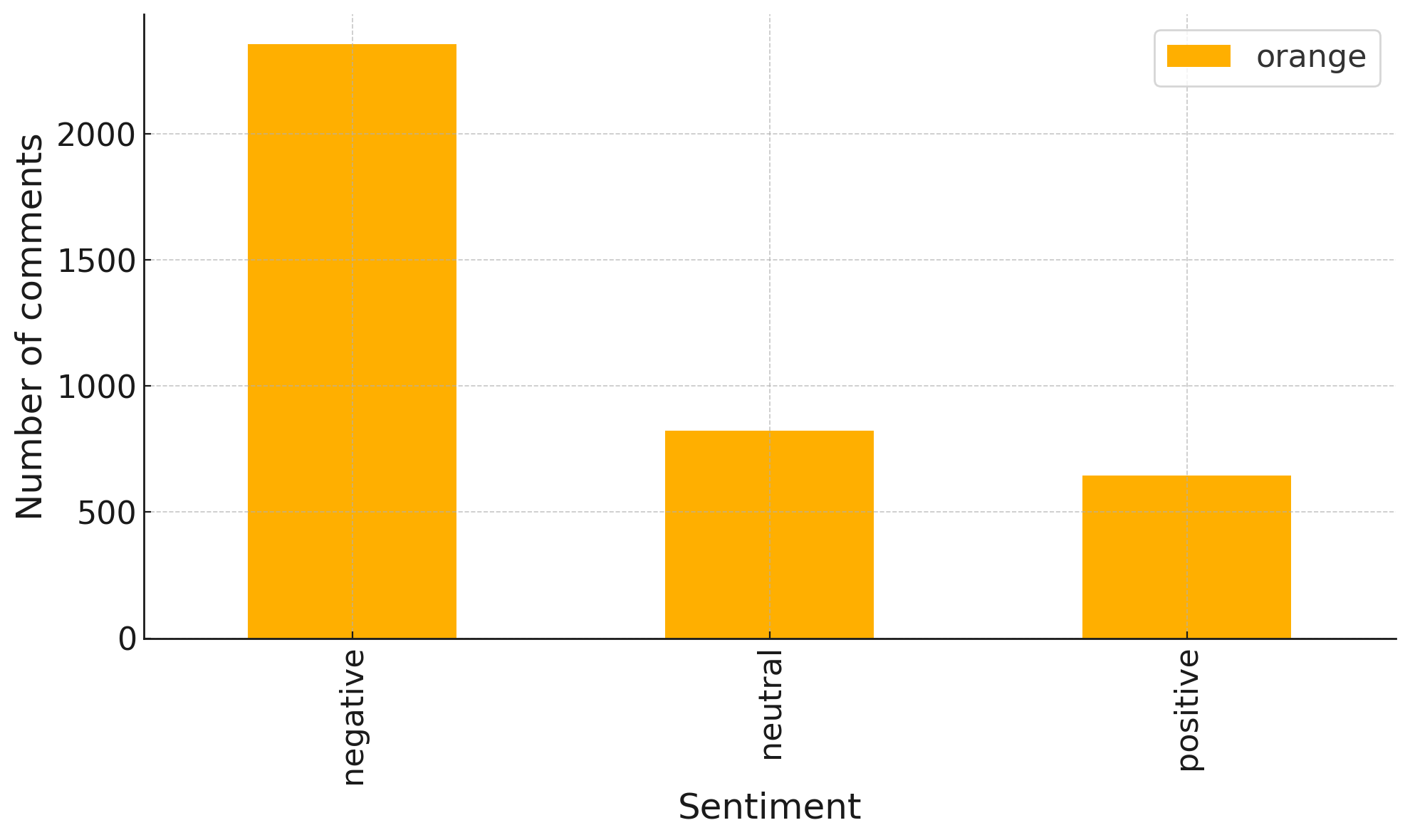}}
\caption{Distribution of user sentiment towards Internet packages and network of the Orange operator.}
\label{orange}
\end{figure}

\begin{figure}[htbp]
\centerline{\includegraphics[width=1\textwidth, keepaspectratio]{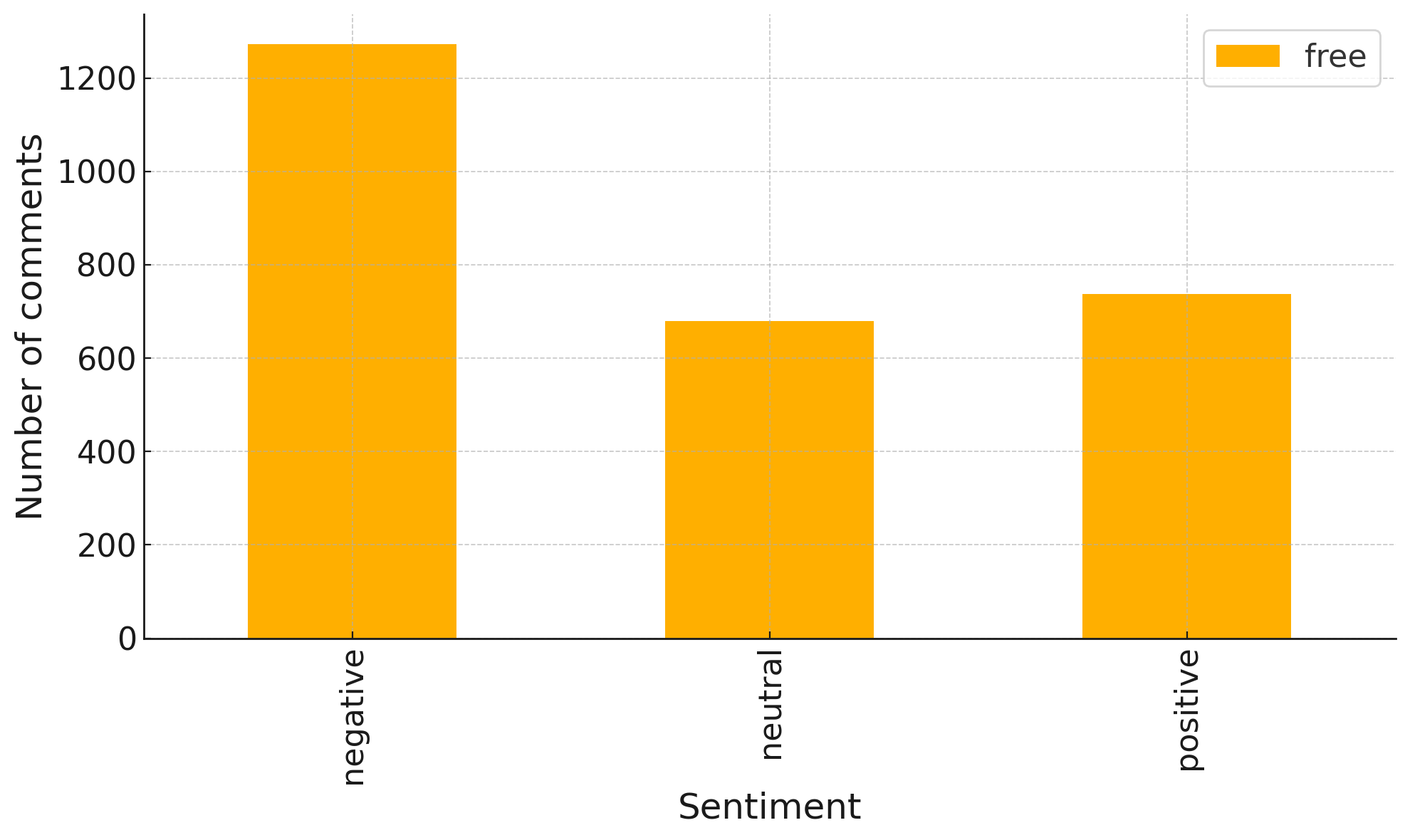}}
\caption{Distribution of user sentiment towards Internet packages and network of the Free operator.}
\label{free}
\end{figure}

\begin{figure}[htbp]
\centerline{\includegraphics[width=1\textwidth, keepaspectratio]{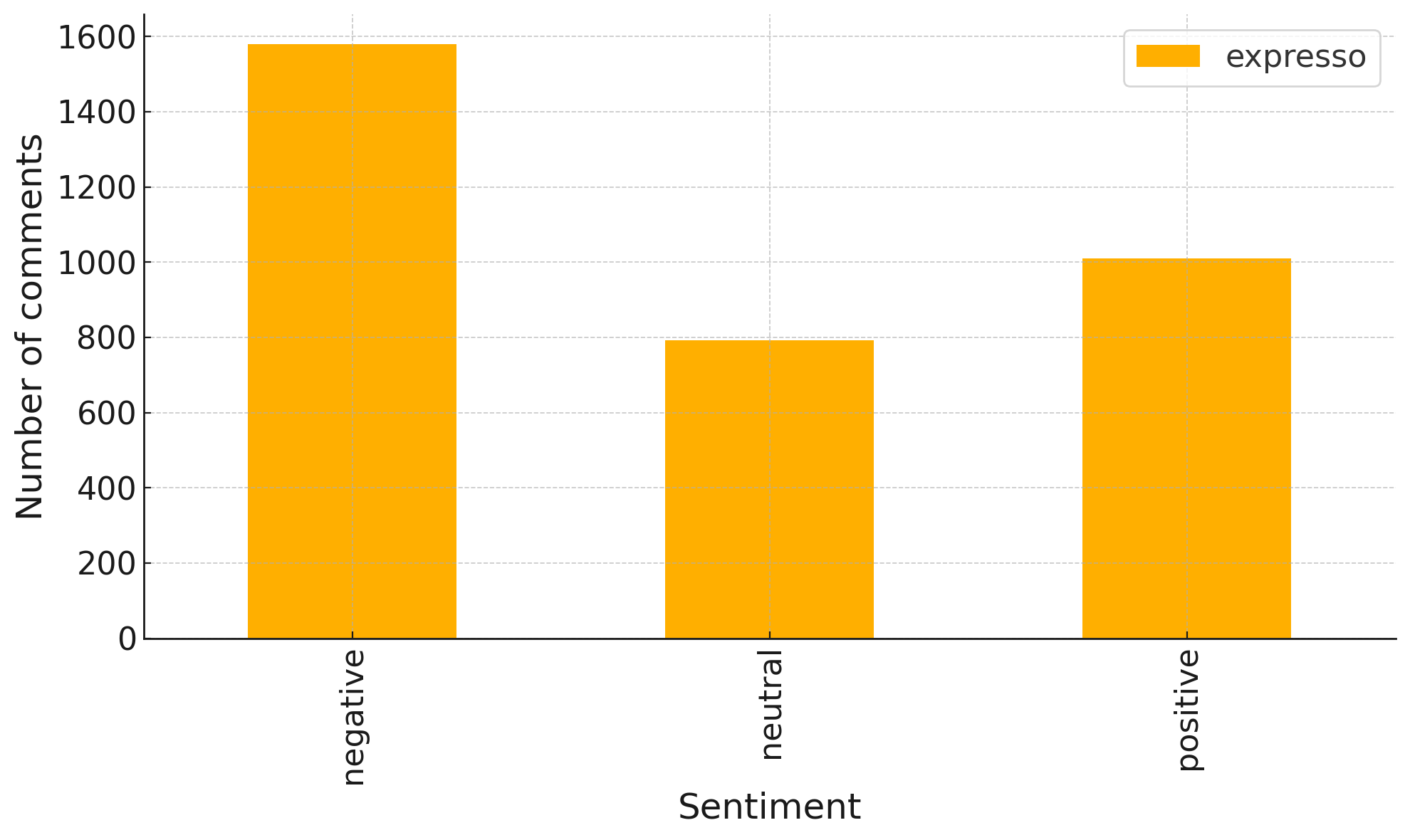}}
\caption{Distribution of user sentiment towards Internet packages and network of the Expresso operator.}
\label{expresso}
\end{figure}

\begin{figure}[htbp]
\centerline{\includegraphics[width=1\textwidth, keepaspectratio]{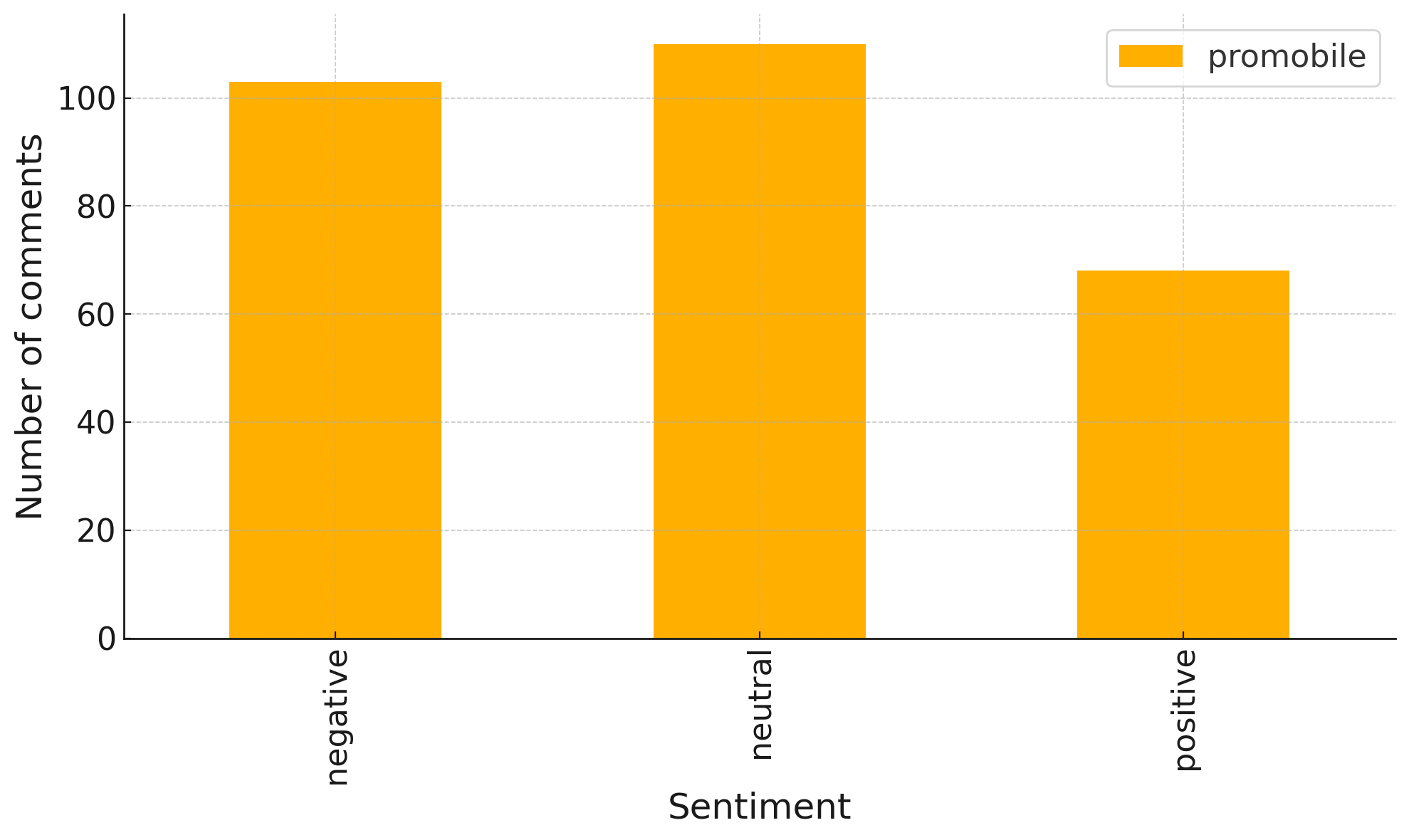}}
\caption{Distribution of user sentiment towards Internet packages and network of the Promobile operator.}
\label{promobile}
\end{figure}

Sentiment analysis is the process of analyzing digital text to determine if the emotional tone of the message is positive, negative, or neutral. To perform this task, we first used the GPT4o model \cite{openai2024gpt4ocard}, which is OpenAI's state-of-the-art Large Language Model. It is a multimodal model with text, visual and audio input and output capabilities, building on the previous iteration of OpenAI’s GPT-4 with Vision model, GPT-4 Turbo. However, the capabilities of this model on low-resource languages like Wolof are limited, although outperforming open-source alternatives as studied in \cite{adelani2024irokobenchnewbenchmarkafrican}. The preliminary tests we carried out on our Wolof data highlighted the limitations of this model, which tended to systematically classify the sentiment of Wolof texts as neutral. To remedy this, we used the Google Translate API\footnote{\url{https://cloud.google.com/translate}} to translate all Wolof texts into French before sentient extraction. Google's translation model offers greater robustness to variations in Wolof writing than specialized models like the one presented in \cite{mbayenmt}. Despite this switch to French, which is better supported by GPT4o and multilingual models in general, we observed the same behavior with an overclassification to the neutral sentiment. To mitigate this aspect, we used XLM-T presented in \cite{barbieri2022xlmtmultilinguallanguagemodels}, which is a multilingual model based on XLM-Roberta \cite{xlmr} and pre-trained on nearly 200 million Tweets across some 30 languages (including French). The authors show that a domain-specific model (in this case, social media) is more effective than its general counterpart when it comes to refining task-specific multilingual Language Models.
The sentiments obtained on the Orange data are illustrated in Fig. \ref{orange}, and show a strong negative connotation, as does the overview obtained on the word clouds.
In fact, the price of the packages is generally raised with network problems from time to time, even if in general the network is more stable. We observe a similar trend for Free's data in Fig. \ref{free} and Expresso's in Fig. \ref{expresso}, but with a higher proportion of positive comments.
Subscribers to these two operators are on the whole satisfied with their plans, but more often decry network quality and coverage. A wave of indignation is however increasingly noted over recent package prices suggesting a subtle increase. Promobile shows the highest proportion of positive comments relative to its total number of reviews as illustrated in Fig. \ref{promobile}.
This may be explained by its very recent arrival on the market, and its attempt to maintain attractive prices to attract customers. It is, however, a virtual operator backed by the Orange network, and is nevertheless associated with it in the eyes of users which is not much appreciated by them. This analysis points to a general dissatisfaction with operators on the part of users, as the trade-off between network quality and affordability is difficult for operators to satisfy. Better regulation of the local telecoms market and support mechanisms from the government could lead to a significant drop in costs, as well as greater attractiveness. An entire sector of the economy is developing in Senegal around Internet connectivity, such as e-commerce, delivery, "uberization", e-sport, etc., and the development of these activities is still very much affected by the accessibility of a high-quality network.

\section{Limitation}\label{limit}
Despite the efficiency of our approach to studying data trends, particularly in a low-resource language, we note a few limitations. The over-representation of Facebook data may induce biases, as users may have a different behavior on Twitter. Restrictions on the latter, however, make it difficult to collect a substantial data on this platform to balance the final corpus. Our approach also relies mainly on the translate-test principle, which involves translating data from a source language to a target one, in order to use tools or approaches that work better in the target language. This method has proved highly effective in low-resource environments as studied in \cite{chen2024translationfusionimproveszeroshot}, but errors in the translation process tend to propagate to the subsequent steps, potentially inducing additional bias.

\section{Conclusion}\label{concl}
In this paper, we studied the sentiment of Senegalese users towards the cost of accessing Internet services from established operators. We collected a substantial corpus on Twitter and Facebook, the latter being the network with the highest concentration of users in Senegal. After an initial phase of data pre-processing, we studied the trends emerging from the most frequently used words in the comments. We took this analysis a step further by highlighting the sentiments expressed in these posts, underlining a general dissatisfaction with the quality/price ratio of the various operators' offers. The overall study is, however, subject to a number of biases relating to data balance, which will need to be strengthened through further data collection. We also intend to carry out an annotation of the final corpus in order to build up an open sentiment analysis dataset for the Wolof language. This will ultimately provide a more suitable and therefore more powerful classification model to facilitate further studies on a variety of topics.

\section*{Acknowledgment}

This project has been funded by Fondation Botnar.

%
%
%
%
\bibliographystyle{spbasic_unsrt}
\bibliography{mybibliography}
\end{document}